\documentclass[sigconf]{acmart}
\usepackage{subfiles}

\AtBeginDocument{%
  }

\setcopyright{acmlicensed}
\copyrightyear{2018}
\acmYear{2018}
\acmDOI{XXXXXXX.XXXXXXX}

\acmConference[Conference acronym 'XX]{Make sure to enter the correct
  conference title from your rights confirmation emai}{June 03--05,
  2018}{Woodstock, NY}
\acmISBN{978-1-4503-XXXX-X/18/06}

\begin{document}

\title{3M: Multi-modal Multi-task Multi-teacher Learning for Game
Event Detection}

\author{Thye Shan Ng}
\email{shan.ng@uwa.edu.au}
\affiliation{
    \institution{University of Western Australia}
    \country{Australia}
}
\author{Feiqi Cao}
\email{fcao0492@uni.sydney.edu.au}
\affiliation{
    \institution{University of Sydney}
    \country{Australia}
}

\author{Soyeon Caren Han}
\email{caren.han@unimelb.edu.au}
\affiliation{
    \institution{University of Melbourne}
    \country{Australia}
}
% \affiliation{%

\renewcommand{\shortauthors}{Ng et al.}

\begin{abstract}
Esports has rapidly emerged as a global phenomenon with an ever-expanding audience via platforms, like YouTube. Due to the inherent complexity nature of the game, it is challenging for newcomers to comprehend what the event entails. The chaotic nature of online chat, the fast-paced speech of the game commentator, and the game-specific user interface further compound the difficulty for users in comprehending the gameplay. To overcome these challenges, it is crucial to integrate the Multi-Modal (MM) information from the platform and understand the event. 
The paper introduces a new MM multi-teacher-based game event detection framework, with the ultimate goal of constructing a comprehensive framework that enhances the comprehension of the ongoing game situation.
While conventional MM models typically prioritise aligning MM data through concurrent training towards a unified objective, our framework leverages multiple teachers trained independently on different tasks to accomplish the Game Event Detection. The experiment clearly shows the effectiveness of the proposed MM multi-teacher framework. 
\end{abstract}

% Esports has become a global sensation, drawing immense audiences through livestream platforms such as Twitch and YouTube.
% However, newcomers often struggle to grasp the complexities of the game due to the fast-paced commentary, chaotic chat interactions, and intricate user interfaces.
% To alleviate this, this research introduces a Multi-modal Multi-task Multi-teacher (3M) learning model to enhance viewer comprehension.
% By integrating real-time chat interactions alongside livestream information, the model aims to detect pivotal game events, offering newcomers a clearer understanding of the ongoing game. 
% This study processes diverse streams of information from the multifaceted platform to enhance the viewer experience, ultimately contributing to the advancement of esports engagement.



%   \includegraphics[width=\textwidth,height=5.5cm]{image/eventchunk1.png}
%   \caption{The segmenting of the continous livestream data into discrete data chunks via livestream audio by employing OpenAI's Whisper model}
%   \label{fig:eventchunk}

\received{20 February 2007}
\received[revised]{12 March 2009}
\received[accepted]{5 June 2009}

\maketitle

\section{Background}

The advent of online streaming platforms\footnote{For example, Twitch. \nolinkurl{https://twitch.tv/}} gave rise to a new era for gaming channels, providing enthusiasts and audiences with the opportunity to spectate gaming events in real-time from the comfort of their homes. 
Much akin to live broadcasts of traditional sports events on television, esports competitions are also presented through live streaming. 
However, what distinguishes gaming broadcasting platforms from traditional sports is the integration of a chat function, fostering a much more interactive viewing experience. 
Audiences are empowered to express their real-time reactions to the gameplay, offering an immediate outlet for emotions ranging from excitement to frustration. 

For long-time fans of esports who are well-versed in the game's narrative, comprehending the overall game situation comes naturally.
However, while esports livestreams provide an immersive viewing experience, they present challenges for newcomers to comprehend the fast-paced gameplay and understand the overall game situation.
Firstly, the game itself is characterised by its high speed and dynamic nature, making it difficult for newcomers to keep up with the action. 
Secondly, while commentators are present to provide an explanation of the game situation \cite{ishigaki-etal-2021-generating}, their speech is often delivered rapidly compared to traditional sports broadcasting. 
Lastly, the chat function accompanying the livestream is typically filled with clutter \cite{nematzadeh2019information}, including numerous misspellings and abbreviations, further complicating the understanding of the event. 
Thus, in order to allow for a better comprehension of the overall game situation, it becomes imperative to address the issues present within each aspect of the livestream platform.
%------

For the gameplay livestream, previous works to enhance the game comprehension include usage of additional visualisation tools via an additional smartphone medium \cite{kokkinakis2020dax}, or through additional overlays on the livestream screen to convey the notable performance of players \cite{block2018narrative}. 
Furthermore, there are also efforts focused on generating commentary by leveraging in-game data and video content using various deep learning models, ranging from seq2seq \cite{ishigaki-etal-2021-generating} to Transformers \cite{zhang-etal-2022-moba,tanaka-esports,wang2022esports}.
These approaches typically rely on data sources extracted solely from the game itself, utilising a single modality.

On the other hand, there exist investigations addressing the comprehension of livestream chat. 
Notable endeavours in this domain encompass the detection of toxic chat \cite{weld2021conda,kim2022understanding}, as well as examinations into the characteristics of conversational language \cite{ford2017chat}.
These explorations specifically delve into how the language in livestream chat deviates from conventional language grammar.

While prior research has contributed to an enhanced comprehension of both gaming dynamics and chat platforms, there remains a gap in the literature concerning the integration of livestream content and chat platforms together to establish a comprehensive understanding of the overall gaming situation. 
Our method, therefore combines the multiple modalities, sourcing from both the game and audience reaction from the livestream platform to facilitate a framework for game event detection. 

%----
By addressing the above-mentioned challenges, we contribute to the gaming and NLP community in the following ways:

\begin{itemize}

    \item Propose a 3M game event detection framework that leverages game-specific knowledge to comprehend the actions unfolding within the gameplay. This framework incorporates various teachers, including chat teacher, caster transcript teacher, and game audio teacher, each fine-tuned on tasks specific to its respective modalities.

    \item Demonstrate the significance of incorporating teacher expertise trained independently on different tasks, in contrast to traditional MM models that follow a unified training objective, resulting in a more robust system.
    
    % Evaluate the effectiveness of game comprehension with the inclusion or exclusion of specific modalities. This comparative analysis offers insights into the individual contributions of each modality towards the overall comprehension output.
    
\end{itemize}

% \section{Related Work}
%     \subfile{chapters/relworks}
    
\section{Method}
    \begin{figure}
        \centering
        \includegraphics[width=\linewidth]{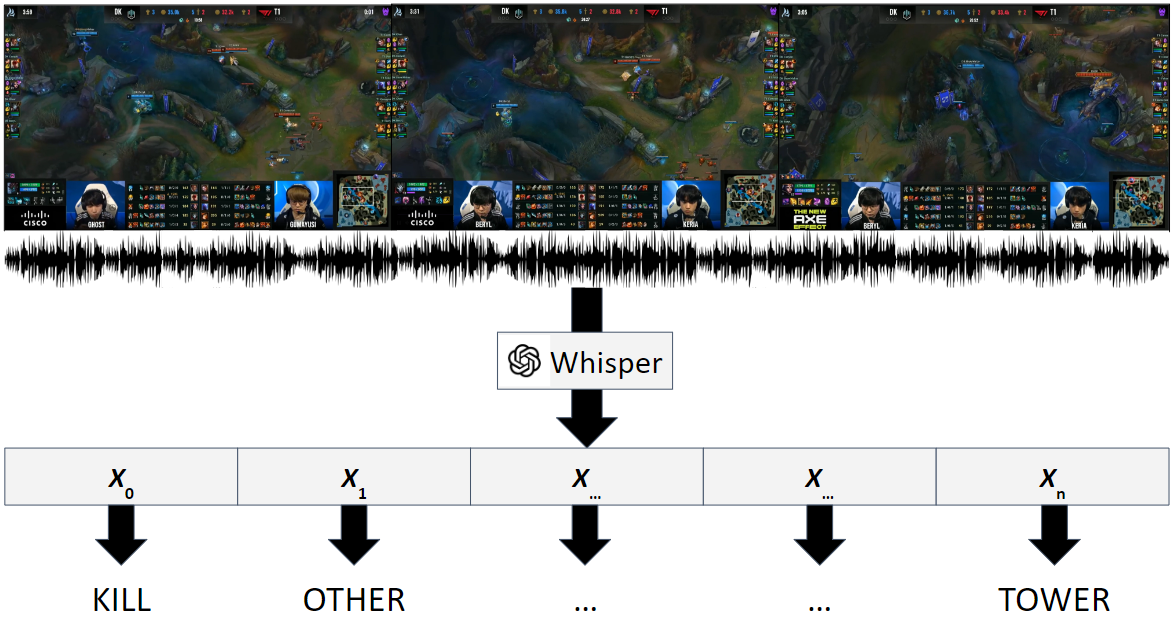}
        \caption{The segmenting of the continous livestream data into discrete data chunks via livestream audio by employing OpenAI's Whisper model. Where each data chunk $X_i$ consist of a game event action as our $Y$}
        \label{fig:eventchunk}
\end{figure}

\subsection{Data Processing}
Due to the continuous nature of the livestream platform, we begin by segmenting the data into discrete chunks. 
To accomplish this, we employ OpenAI's Whisper model \cite{radford2023robust}.
This model transcribes the raw audio content of the broadcast into transcripts, which generate distinct windows as a byproduct, as shown in Figure \ref{fig:eventchunk}.
Following this, we employ the defined window size to split all three modalities accordingly. 
The three modalities are chat, audio, and transcript.
For each of the segmented data $X_i$, the ground truth $Y_i$ is defined as the pivotal actions that occurred within the gameplay, they are:
\begin{itemize}
    \item \emph{KILL} - when player A defeated player B
    \item \emph{DRAGON} - when a dragon entity is defeated by a player
    \item \emph{TOWER} - when a tower is destroyed by a player
    \item \emph{OTHER} - when no significant actions have taken place
\end{itemize}

\subsection{3M Fine-Tuning}
To imbue each of the teacher models with game-specific expertise knowledge, we fine-tune each of the teachers under different modalities and settings. 
Specifically, we introduce $N=3$ teachers, which include the audio teacher, the chat teacher, and the transcript teacher. 
All teacher models are separately fine-tuned to different tasks, covering the various aspects of the livestream platform.

\textbf{Audio-based Action Detection Teacher}
For the audio teacher, we initialise a $K=12$ layer Audio Spectrogram Transformer (AST) pretrained model \cite{gong2021ast}.
Subsequently, we undertake a fine-tuning process to adapt this model for the purpose of identifying noteworthy events occurring during LoL match.
The input data for this model is specified as the unprocessed audio extracted from the livestream of the gameplay. 
In the course of fine-tuning, we employ a straightforward cross-entropy loss function to optimise the model's performance. 

\textbf{Chat-based Emotion Tagging Teacher}
For the chat teacher, we initialise a $K$ layer XLM-RoBERTa \cite{conneau2020unsupervised} pretrained model and fine-tuned for the specific task of discerning token-level emojis and emotes within chat interactions. 
The input for this model is specified as the raw chat data extracted from the chat section of the livestream.  
For fine-tuning, we utilise a simple cross-entropy loss function to optimise the model's performance.

\textbf{Transcript-based Game Entity Tagging Teacher}
For the transcript teacher, we initialise a $K$ layer RoBERTa \cite{liu2019roberta} pretrained model and fine-tuned it to identify token-level LoL-specific terminologies within the transcript.
The input for this model is specified as the transcript of the commentary, which is converted from raw audio to text via OpenAI's Whisper model \cite{radford2023robust}.
To fine-tune this, a simple cross-entropy is utilised.

\subsection{3M Knowledge Distillation}
\begin{figure}
    \centering
    \includegraphics[scale=0.3]{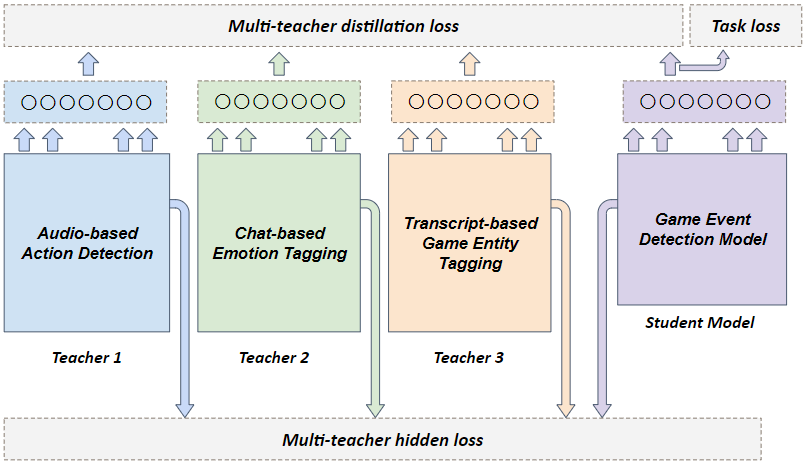}
    \caption{Proposed 3M Distillation Framework, where the student receives knowledge from 3 teachers, each containing different game expertise}
    \label{fig:multitaskmodel}
\end{figure}
Next, we introduce our proposed 3M framework, which is shown in Figure \ref{fig:multitaskmodel}.
Drawing inspiration from architecture proposed by Wu et al. \cite{wu2021one}, we use two loss functions for knowledge distillation.
They are multi-teacher hidden loss and multi-teacher distillation loss.

The multi-teacher hidden loss aims to transfer knowledge of the teacher's hidden states to the student model \cite{sun2019patient}. 
In the framework, There are $N$ number of fine-tuned teacher models, and all of them have $K$ Transformer layers. 
Once a smaller student model of $M=8$ Transformer layers is initialised, each teacher's layer is then compared with its corresponding student's layer.
Each $i$th student hidden layer, denoted as $H^S_i$ is directly aligned with a corresponding $i$th teacher layer, denoted as $H^T_i$.
For the remaining layers of the teacher hidden states, a naive approach is utilised.
Specifically, we directly map the remaining $H^T_j$ layers with the final student layer $H^S_M$.
Following that, a mean squared error (MSE) is applied to encourage the student model to have similar functions to teacher models. 
The hidden loss, therefore, $\mathcal{L}_{MT-Hid}$ is as follows:
\begin{equation}\label{equation:multi_teacher_hidden_loss}
    \mathcal{L}_{MT-Hid} = \sum_{t=1}^N(\sum_{i=1}^K\text{MSE}(H_i^S,W_{ti}H_i^T) +\sum_{j=K}^M\text{MSE}(H_K^S, W_{tj}H_j^T))
\end{equation}
where $W_{ti}$ is a trainable transformation matrix. 
Note that since the teacher models are fine-tuned to their game expertise task, tampering the stored expertise information within this model is not ideal.
Hence, all hidden layers $H^T$ of the teacher will be frozen when performing knowledge distillation. 

The primary objective of the multi-teacher distillation loss is the transmission of knowledge encapsulated within the soft labels, originating from multiple teacher models, to the student model.
In our design, we replace the final layer of the teacher models for task-specific predictions with a trainable output layer, which is of the same dimension as the student output layer.
Given the specific fine-tuning of each teacher model to their respective game expertise task, we do not anticipate precise game situation classification accuracy from the teacher model in this context. 
Instead, we aim to distill modality-related information from the ensemble of teacher models to the student. 
The multi-teacher distillation loss $\mathcal{L}_{MT-Dis}$ is formulated as follows:
\begin{equation}\label{equation:multi_teacher_distillation_loss}
    \mathcal{L}_{MT-Dis} = \sum_{i=1}^N \frac{\text{CE}(y_i,y_s)}{1 + \text{CE}(y,y_i)}
\end{equation}
where the output logits of the teacher model are compared with the student model, as well as the ground truth label via cross entropy.

Finally, we incorporate ground truth labels to compute the task-specific loss $\mathcal{L}_{Task}$ based on the predictions of the student model for game event detection using transcripts as the input for the student model.
\begin{equation}\label{equation:student_task_loss}
    \mathcal{L}_{Task} = \text{CE}(y,y_s)
\end{equation}
The final loss function $\mathcal{L}$ is the summation of multi-teacher hidden loss, multi-teacher distillation loss and task-specific loss, which is formulated as:
\begin{equation}\label{equation:total_loss}
    \mathcal{L} = \mathcal{L}_{MT-Hid} + \mathcal{L}_{MT-Dis} + \mathcal{L}_{Task}
\end{equation}

\section{Experiments}
    \begin{table}[]
    \centering
    \begin{tabular}{lcccc}
        \hline
        \textbf{Label}  & \textbf{Train} & \textbf{Test}  \\
        \hline
        KILL &  2647 & 164 \\
        DRAGON & 1659 & 76 \\
        TOWER & 1233 &  59 \\
        % KILL DRAGON & 321 & 1.17 & 14 & 0.98\\
        % KILL TOWER & 231 & 0.84 & 10 & 0.70\\
        % DRAGON TOWER & 83 & 0.30 & 4 & 0.28\\
        % KILL DRAGON TOWER & 10 & 0.04 & 0 & 0.0\\
        OTHER & 21357 & 970 \\
        \hline
        Total & 26896 & 1249
    \end{tabular}
    \caption{Data distribution of game events across a total of 45 distinct esports tournaments, where \textbf{Other} is defined as intervals within the game characterised by the absence of event actions}
    \label{tab:datasize}
\end{table}

\begin{table*}[]
    \centering
    \begin{tabular}{l|ccc|ccc|ccc|ccc}
        \hline
        & \multicolumn{6}{c|}{Game-MUG} & \multicolumn{6}{c}{3M} \\
        \hline
        & \multicolumn{3}{c|}{OTHER $\circ$} & \multicolumn{3}{c}{OTHER $\times$} & \multicolumn{3}{|c|}{OTHER $\circ$} & \multicolumn{3}{|c}{OTHER $\times$} \\ 
        \hline 
        \textbf{Label}  & \textbf{P} & \textbf{R} & \textbf{F1} & \textbf{P} & \textbf{R} & \textbf{F1} & \textbf{P} & \textbf{R} & \textbf{F1} & \textbf{P} & \textbf{R} & \textbf{F1}\\
        \hline
         KILL & 0 & 0 & 0 & 0.859 & 0.421 & 0.565 & 0.646 & 0.498 & 0.562 & 0.762 & 0.781 & 0.772  \\
         DRAGON & 0 & 0 & 0 & 0.160 & 0.286 & 0.205 & 0.408 & 0.233 & 0.297 & 0.763 & 0.468 & 0.580\\
         TOWER & 0 & 0 & 0 & 0 & 0 & 0 & 0.220 & 0.092 & 0.130 & 0.390 & 0.535 & 0.451\\
         OTHER & 0.740 & 1 & 0.851 & - & - & - & 0.705 & 0.859 & 0.775 & - & - & - \\
         \hline
         AVG$_\text{MACRO}$ & 0.340 & 0.236 & 0.257 & 0.185 & 0.250 & 0.213 & 0.495 & 0.421 & 0.441 & 0.638 & 0.595 & 0.601
         
    \end{tabular}
    \caption{(P)recision, (R)ecall, and (F1) of game event detection model from our proposed 3M Framework, as well as the baseline model Game-MUG. Both models are trained with dataset that includes the OTHER label, as well as dataset that excludes the OTHER label}
    \label{tab:3m_baseline}
\end{table*}

\begin{table*}[]
    \centering
    \begin{tabular}{ccccccc|c}
    \hline
        Audio & Chat & Transcript & KILL & TOWER & DRAGON & OTHER & AVG$_\text{MACRO}$\\
    \hline
        $\circ$ & $\times$ & $\circ$ & 0.409 & \textbf{0.724} & 0.034 & 0.131 & 0.325\\
        $\times$ & $\circ$ & $\circ$ & 0.628 & 0.355 & 0.136 & \textbf{0.812} & 0.483\\
        $\circ$ & $\circ$ & $\times$ & \textbf{0.804} & 0.237 & 0.153 & 0.595 & 0.447\\
        $\circ$ & $\circ$ & $\circ$ & 0.646 & 0.408 & \textbf{0.220} & 0.705 & \textbf{0.495}\\
    \hline
    \end{tabular}
    \caption{The impact on performance (precision) following the integration of various combinations of teacher expertise model}
    \label{tab:2mloss}
\end{table*}

\subsection{Datasets and Evaluation Metrics}
The game domain-specific dataset employed in our experiment is curated from LoL matches broadcasted on the platforms Twitch and YouTube, of which three aspects are featured in the dataset: game match event logs, audio data and textual discussions.
The dataset spans across 45 distinct esports tournaments, comprising a total of 216 unique LoL matches \cite{zhang2024gamemug}.
Employing the segmentation strategy detailed in Section 2.1, these matches are further decomposed into 28,145 individual data instances, subsequently partitioned into a 95/5 train test split.
The total number of instances for each event label is as described in Table \ref{tab:datasize}, where OTHER refers to when there are no particular actions or events happening.

To evaluate the performance of our proposed 3M framework, the metrics that we use are as follows: 1) Precision-Score, 2) Recall-Score and 3) F1-Score. 

\subsection{Training Details}
In our experiments, the default value of $K=12$ attention layers is instantiated for AST (audio teacher) \cite{gong2021ast}, XLM-RoBERTa (chat teacher) \cite{conneau2020unsupervised} and RoBERTa  (transcript teacher) \cite{liu2019roberta} models to distill an $M=8$ layer student model.
For the student model, we use the first $M$ layer of RoBERTa to initialise the parameters for our framework.
The learning rate for fine-tuning the teacher model is set to (1e-5, 1e-7) while the learning rate for performing 3M knowledge distillation is set to (1e-4, 1e-7) using the cyclical approach \cite{smith2017cyclical}.
The optimiser used is AdamW \cite{DBLP:journals/corr/abs-1711-05101}, and a dropout rate of 0.1 is consistently applied to all teacher and student models throughout the training of the model.
Finally, all implementations are based on PyTorch \cite{paszke2019pytorch} and HuggingFace \cite{wolf2020transformers}.

\textbf{Baseline.}
We evaluate our proposed 3M model by comparing its performance against the baseline model Game-MUG BERT \cite{zhang2024gamemug}, employing the dataset segmented through our proposed segmentation strategy. 
The focus of this comparative analysis lies in the game situation understanding module within the Game-MUG framework. 

\subsection{Results}
Table \ref{tab:3m_baseline} outlines the overall performance of our proposed 3M framework and the baseline Game-MUG \cite{zhang2024gamemug}. 
Both models are trained with dataset which includes the OTHER label, as well as dataset without the OTHER label. 
When considering the macro averages, the 3M framework demonstrates a notably superior performance.

We note that the Game-MUG baseline, designed for classifying game events without the presence of the OTHER label, struggles to identify game events when the OTHER label is introduced.
In this scenario, all of the evaluation metrics received a score of 0, apart from OTHER.
This indicates that all predictions made by the baseline incorrectly classified events as OTHER.
Furthermore, while the baseline model is able to achieve better classification scores under the absence of OTHER, our 3M framework still consistently outperforms. 
We hypothesise that this is likely due to the class imbalance, as well as the aggregation of all modalities as a single input.
Furthermore, we hypothesise that the improved classification score achieved by our 3M framework is likely attributed to the separated processing of modalities, offering enhanced discrimination.

We then evaluate the performance of our proposed 3M framework on the testing set, with supervision from all three teacher models, as illustrated in Table \ref{tab:3m_baseline}.
We observe that while OTHER has achieved the highest classification score, the KILL label follows as the second highest, followed by DRAGON and TOWER.
Additionally, the accuracy of the labels is directly associated with the dataset distribution, indicating that more frequent data corresponds to higher classification scores, even with the weighing of samples.
However, when training our 3M framework by removing OTHER label from the dataset, there is notable increase in performance across all events. 
For instance, while the precision score of TOWER is 0.220, its performance went up to 0.390 when trained without OTHER. 

\textbf{Effect of multiple teachers}
To further analyse the contributions of each game expertise teacher towards the student, we conduct an ablation study of the performance in a scenario involving only two teachers, and compare them with the full ensemble. 
The precision scores are as illustrated in Table \ref{tab:2mloss}.
In the table, $\times$ denotes teacher models not in use, while $\circ$ signifies the model in use for the distillation process.
From the result, it is evident that the model's proficiency in the classification task varies depending on the teacher combination. 
For instance, Audio and Chat teacher combination excels in TOWER, while Chat and Transcript teacher combination excels in performance for the OTHER label.
Even though the performance of 3M (with all modalities) is lower when compared label by label, its average precision score is notably higher.
We hypothesise that while the addition of teacher may enhance the overall average performance, it also may have introduced noise in separate modalities, leading to a lower performance for specific labels. 

Lastly, while the performance of 3M for the labels KILL and OTHER shows high accuracy, its classification accuracy for TOWER and DRAGON events remains relatively low.
KILL events likely receive more attention from the audience and commentators, providing clearer cues in the data. 
In contrast, TOWER and DRAGON events, while important in gameplay, may not provoke as much reaction, resulting in fewer discernible features for the model to learn from.
Regardless, the suboptimal performance remains a pertinent avenue for further research. 

\section{Conclusion}
    In this paper, we introduce the Multi-teacher Multi-task Multi-teacher framework named 3M for game event detection, which imbues the student model with strong game knowledge from multiple teacher game expertise models via knowledge distillation.
By learning from diverse task-specific experts, which are the teacher models, the student model was able to capture a broader range of game-related features and nuances, enhancing its performance in event detection tasks.
Our study showcases the capabilities of the framework, offering a novel approach to leverage diverse game expertise for improved performance.

% \section{Rights Information}

% Authors of any work published by ACM will need to complete a rights
% form. Depending on the kind of work, and the rights management choice
% made by the author, this may be copyright transfer, permission,
% license, or an OA (open access) agreement.

% Regardless of the rights management choice, the author will receive a
% copy of the completed rights form once it has been submitted. This
% form contains \LaTeX\ commands that must be copied into the source
% document. When the document source is compiled, these commands and
% their parameters add formatted text to several areas of the final
% document:
% \begin{itemize}
% \item the ``ACM Reference Format'' text on the first page.
% \item the ``rights management'' text on the first page.
% \item the conference information in the page header(s).
% \end{itemize}

% Rights information is unique to the work; if you are preparing several
% works for an event, make sure to use the correct set of commands with
% each of the works.

% The ACM Reference Format text is required for all articles over one
% page in length, and is optional for one-page articles (abstracts).

% \begin{acks}
% \end{acks}

\bibliographystyle{ACM-Reference-Format}
\bibliography{main}

\appendix









\end{document}